\documentclass[review,3pt]{elsarticle}

\usepackage{lineno,hyperref}

\usepackage{amssymb}
\usepackage{amsmath}
\usepackage{amssymb}
\usepackage{amsfonts}
\usepackage{booktabs}
\usepackage{multirow}
\usepackage{multicol}
\usepackage{verbatim}
\usepackage{color}
\usepackage{algorithm}
\usepackage{algorithmicx}
\usepackage{algpseudocode}
\usepackage{graphicx}

\usepackage{enumerate}

\makeatletter
\def\hlinew#1{%
	\noalign{\ifnum0=`}\fi\hrule \@height #1 \futurelet
	\reserved@a\@xhline}
\makeatother

\begin{document}

\begin{frontmatter}



\title{Line Drawing Guided Progressive Inpainting of Mural Damage}


\author[1,2,3]{Luxi Li}
\ead{lucylee@whu.edu.cn}

\author[2,6]{Qin Zou \corref{cor1}}
\ead{qzou@whu.edu.cn}

\author[3,6]{Fan Zhang} 
\ead{zhangfan@whu.edu.cn}

\author[4]{Hongkai Yu} 
\ead{hongkaiyu2012@gmail.com}

\author[1,5]{Long Chen} 
\ead{long.chen@ia.ac.cn}

\author[2]{Chengfang Song} 
\ead{songchf@whu.edu.cn}

\author[3,6]{Xianfeng Huang} 
\ead{huangxf@whu.edu.cn}

\author[6]{Xiaoguang Wang} 
\ead{wxguang@whu.edu.cn}

\author[3,7]{Qingquan Li} 
\ead{liqq@szu.edu.cn}

\address[1]{Department of Computer Science Technology, United International College of Beijing Normal University - HongKong Baptist University, Zhuhai, China}
\address[2]{Machine Vision and Robotics Laboratory, School of Computer Science, Wuhan University, Wuhan, China}
\address[3]{State Key Laboratory of Surveying, Mapping, and Remote Sensing Information Engineering, Wuhan University, Wuhan, China}
\address[4]{Department of Electrical Engineering and Computer Science, Cleveland State University, OH, USA}
\address[5]{Institute of Automation, Chinese Academy of Sciences, Beijing, China}
\address[6]{Cultural Heritage Intelligent Computing Laboratory, Wuhan University, Wuhan, China}
\address[7]{Guangming Laboratory, Shenzhen University, Shenzhen, China}


\cortext[cor1]{Corresponding author}

\begin{abstract}
Mural image inpainting is far less explored compared to its natural image counterpart and remains largely unsolved. Most existing image-inpainting methods tend to take the target image as the only input and directly repair the damage to generate a visually plausible result. These methods obtain high performance in restoration or completion of some pre-defined objects, e.g., human face, fabric texture, and printed texts, etc., however, are not suitable for repairing murals with varying subjects and large damaged areas. Moreover, due to discrete colors in paints, mural inpainting may suffer from apparent color bias. 
To this end, in this paper, we propose a line drawing guided progressive mural inpainting method. It divides the inpainting process into two steps: structure reconstruction and color correction, implemented by a structure reconstruction network (SRN) and a color correction network (CCN), respectively. In structure reconstruction, SRN utilizes the line drawing as an assistant to achieve large-scale content authenticity and structural stability. In color correction, CCN operates a local color adjustment for missing pixels which reduces the negative effects of color bias and edge jumping. The proposed approach is evaluated against the current state-of-the-art image inpainting methods. Qualitative and quantitative results demonstrate the superiority of the proposed method in mural image inpainting. The codes and data are available at {https://github.com/qinnzou/mural-image-inpainting}\ .
\end{abstract}

\begin{keyword}
	Mural inpainting \sep cultural heritage \sep mural image dataset \sep image inpainting \sep image restoration




\end{keyword}

\end{frontmatter}


\section{Introduction}\label{sec:introduction}

As one of the most cherished world cultural heritages, murals in Dunhuang Mogao Grottoes represent more than one thousand years of Chinese history and culture from the late Jin Dynasty to the early Song Dynasty~\cite{li2018dating}. In this long history, due to natural weathering and human damage, Mogao Grottoes suffered from  cracking, color fading, and some other forms of distresses, which gradually destroyed the murals. Traditional mural restoration methods physically repair the cultural relics, making the restoration result irreversible, which may cause secondary damage to the murals~\cite{wang2019inpainting,chen2019image}. While digitization technologies allow permanent digital storage of the original shape and appearance, the digital inpainting of murals is of great significance to the preservation of cultural heritage.

\begin{figure}[!t]
	\begin{center}
		\includegraphics[width=1.0\linewidth]{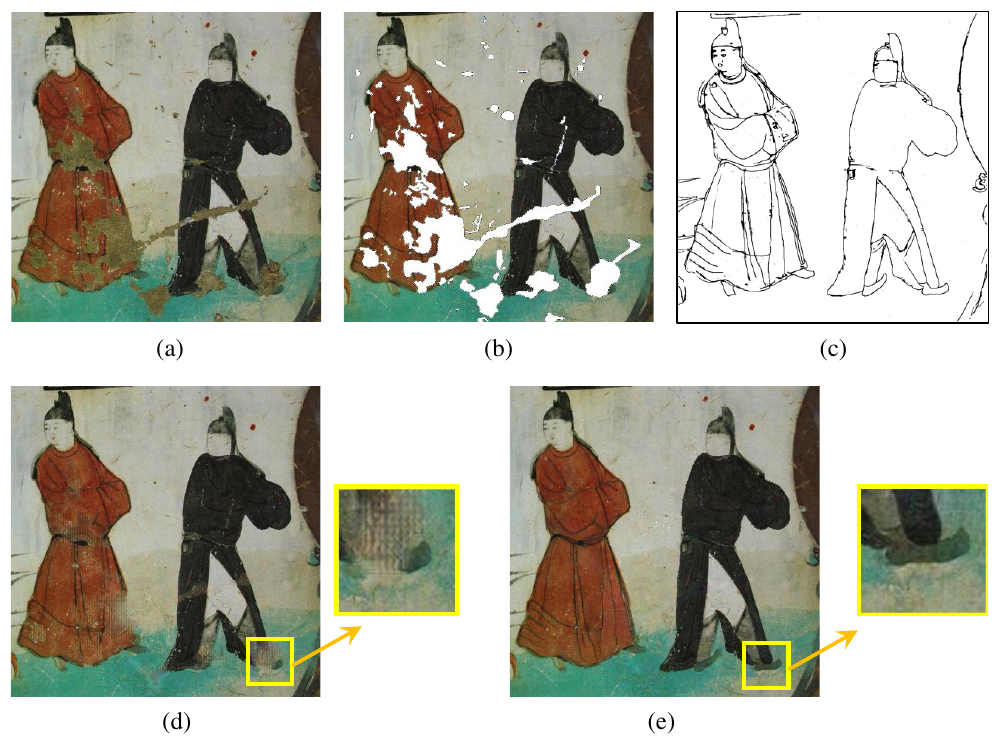}
	\end{center}
	\vspace{-0.1in}
	\caption{Inpainting with and without the assistance of line drawings. (a) An image of damaged mural. (b) Masks on the mural image. (c) Corresponding line drawings of the mural. (d) The inpainting result obtained without line drawings. (e) The inpainting result obtained with line drawings.}
	\label{fig:whylinedrawing}
\end{figure}

\begin{figure}[!t]
	\centering
	\includegraphics[width=0.8\linewidth]{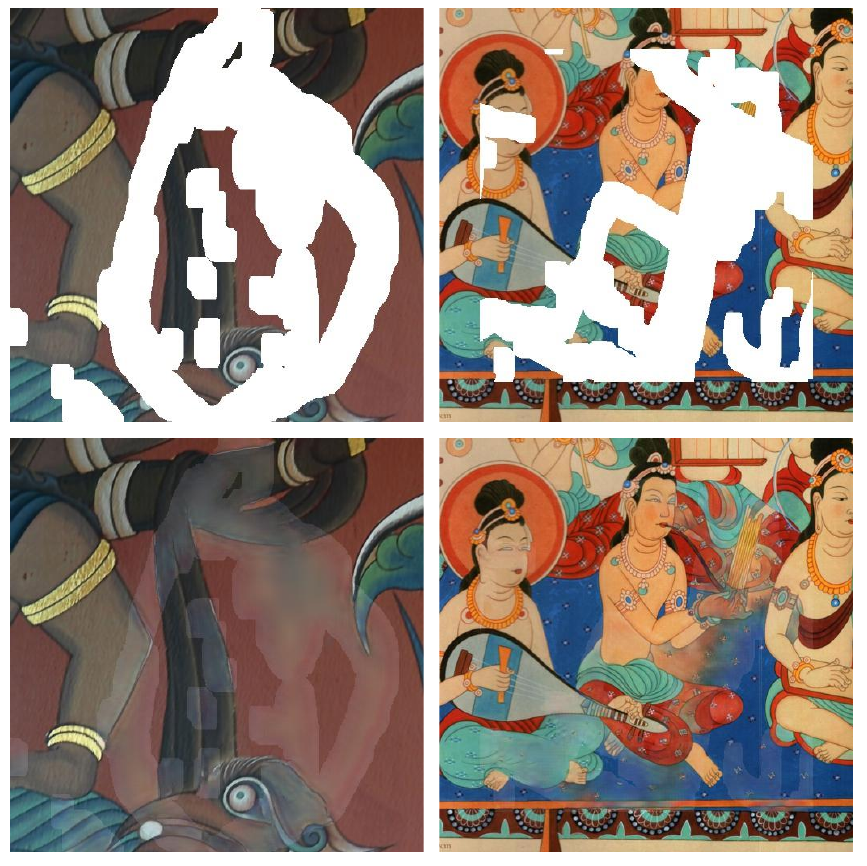}
	\caption{The problem of color bias  in mural inpainting. Top row: mural images with masks in white. Bottom row: the inpainting results. The color bias is apparent between the inside and outside of the mask.}
	\label{fig:colordifference}
\end{figure}

Over the past decade, a number of advanced inpainting methods built on deep convolutional neural networks (DCNN) have been proposed~\cite{pconv,2019FreeForm,zha2022low}. These methods perform well on natural image inpaiting, and produce outstanding results on public available datasets such as CelebA, Places2, and ParisStreetView~\cite{yi2020contextual}. However, they meet problems when applied to mural images. First, the missing parts of the mural may be large and complex, which are difficult to recover directly from the mural images. In mural inpainting, an important requirement is that the murals are restored to the original appearance as much as possible. However, traditional inpainting strategies using only the damaged mural image as input may generate results that do not match the original, even though the results may be visually plausible, as illustrated by Figure~\ref{fig:whylinedrawing}. Second, even with the assistance of line drawings, existing DCNN-based inpainting methods may still suffer from color bias. In most cases, pixels of the whole image are involved in the convolution-pooling computation~\cite{ren2019structureflow}, where position bias of pixels may be introduced in the down- or up-sampling procedure. Meanwhile, the color bias will be apparent when we merge the generated part and the existing part as a lack of color-consistency control in most DCNN-based methods, as illustrated by Figure~\ref{fig:colordifference}.

Although existing methods are not completely suitable for mural inpainting, they do provide us inspirations. The idea of `line first, color next' proposed in~\cite{2019EdgeConnect} predicts the structure lines first, and then combines the lines for structure restoration. It proves that edge maps contribute to structure-aware inpainting. In mural inpainting, the restored images are expected to be consistent with the original murals as much as possible. Considering that edge maps of the mural structure predicted by neural networks are not up to the expert level, we adopt the line drawings manually produced by professional painters as the guidance in the inpaining task. 

In another typical work~\cite{2018Generative}, a coarse-to-fine inpainting strategy was put forward, which divides the inpainting process into two stages. It first produces a rough low-quality inpainted result, and then refines the rough result to obtain a high-quality final result. The strategy achieves a satisfactory result. Accordingly, in our method, we divide the mural inpainting task into two sub-tasks, where one is for structure reconstruction and the other is for color correction. Different from other methods which introduce structure lines and handle structures and colors together~\cite{2019EdgeConnect,2019FreeForm,ren2019structureflow}, we consider the structure and the color separately. The damaged structure is rebuilt in the first inpainting stage, which ignores the color bias. While in the second inpainting stage, the color of the image is adjusted with the completed structures, which would optimize the overall sense, and largely ameliorate the color bias problem.


Based on the above observations and discussions, we propose a progressive mural inpainting method guided by line drawings. The main contributions of this work can be summarized as follows:

\begin{itemize}
	
	\item First, to the problem of large-area mural damage inpainting, we introduce the line drawing as an assistance and the histogram loss as a constraint in the inpainting, which greatly improves the inpainting quality on damaged mural with large holes.
	
	\item Second, to the problem of color bias in mural inpainting, we propose a novel inpainting strategy that decomposes the inpainting into two subtasks -- structure reconstruction and color correction, which not only restores the structure, but also keeps the color consistent.
	
	\item Third, we construct a mural image dataset consisting of 1,714 mural paintings from Dunhuang Mogao Grottoes. Each mural image has a corresponding line drawing. The dataset has been released to the community, which would promote the research of mural inpainting and heritage preservation.
\end{itemize}

The reminder of this paper is organized as follows: Section~\ref{sec:relate} briefly introduces the related work, including the image inpainting and mural inpainting; Section~\ref{sec:method} describes the proposed method in detail; Section~\ref{sec:exp} reports the experiments and results; and Section \ref{sec:conc} concludes the work.

\section{Related Work}\label{sec:relate}


\subsection{Image Inpainting}

Image inpainting has been a hot research topic in the field of computer vision and image processing~\cite{wu2024misl,zhang2023fully,huang2024sparse,yue2023end}. Over the past few years, a number of advanced image inpainting methods built on deep learning techniques have been developed. 

\textbf{Structure-information embedding}. In~\cite{2019EdgeConnect}, Nazeri et al. introduced the concept of 'line first, color next,' conducting separate inpainting procedures for line drawings and color. This approach, recognized as a leading-edge technique, has been extensively compared in various studies. In~\cite{2019Progressive}, a visual structure reconstruction layer was innovatively introduced, which intertwined reconstructions of visual structure and features through parameter sharing, resulting in mutual enhancements. In~\cite{xie2021seamless}, the manga inpainting task was partitioned into structural line inpainting and screen VAE map inpainting, effectively capturing the semantic distinction between structure and screentone. In~\cite{liu2020rethinking}, a mutual encoder-decoder CNN was proposed for the joint recovery of both structures and textures through dual branches. This method effectively eliminates blurring and artifacts arising from inconsistent structural and textural features. In the context of medical images, the integration of structural information into inpainting was explored~\cite{liao2018edge}, significant improved performance was obtained.

\textbf{Coarse-to-fine inpainting strategy}. In~\cite{2018Generative} and \cite{cai2022image}, a coarse result was generated in an initial stage with low resolution. In the subsequent stage, the preceding outcome undergoes enhancement via a refine network. However, these methods cannot handle the inpainting of irregular-shaped regions. In another notable work~\cite{2019Coherent}, a coherent semantic attention layer was integrated into the refinement network to preserve the spatial structure of the image. In a distinct context, Wang et al.~\cite{WANG2022108373} formulated the dynamic-to-static image translation as an image inpainting problem, offering a fresh coarse-to-fine framework. In a different direction, Sagong et al.~\cite{sagong2019pepsi} adopts a decoding network featuring both coarse and inpainting paths, and optimizes the efficiency of convolution operations through the sharing of coarse weights with the inpainting paths. Quan\cite{Refinement} proposes a novel three-stage inpainting frame work with local and global refinement. To improve the efficiency of inpainting, Shin et al.~\cite{Shin2021} propose a novel network architecture called parallel extended-decoder path for semantic inpainting, which consists of a single shared encoding network and parallel decoding networks. It obtained significantly reduced computational cost.

\textbf{Partial convolution}. Partial convolution~\cite{pconv} progressively erodes mask layers to efficiently utilize known information and integrate background details into the gaps. Several methods leverage partial convolution for image inpainting~\cite{yu2023magconv,phutke2023blind}. For instance, in~\cite{2019Progressive}, Li et al. proposed a P-UNet model as the foundational architecture. This innovative approach involves substituting each convolution layer in the U-Net~\cite{ronneberger2015u} with a partial convolution layer, enabling better capture of local information along irregular boundaries. Another instance can be found in~\cite{WANG2020107448}, where partial convolution is employed to counteract the influence of masked areas on the generated outcomes. Moreover, in~\cite{2019FreeForm}, gated convolution techniques were introduced  to enhance the mask mechanism of partial convolution. This shift transforms the rule-based mask approach into a learning-based mechanism, consequently reducing the constraints imposed by the rectangular masks utilized in their earlier work~\cite{2018Generative}. Zhu et al.\cite{zhu2021image} proposed Mask-Aware Dynamic Filtering (MADF) module to effectively learn multi-scale features for missing regions. Wang et al.\cite{wang2021dynamic} introduced the Dynamic Selection Network to distinguish between damaged and valid regions, achieving excellent outcomes. Shang et al.\cite{shang2023deformable} built a novel inpainting method on deformable convolutions,  which can avoid the deformably sampled elements falling into the corrupted regions.

\textbf{Visual Attention}. In \cite{8953485}, Zeng et al. integrated cross-layer attention and pyramid filling mechanisms using a U-Net architecture. This strategy involved training the encoder to grasp region affinity through attention within a high-level semantic feature map, and then transferring this learned attention to its adjacent high-resolution feature map. Another approach, as presented in \cite{WANG2020107448}, introduced a multistage attention module, aimed at achieving superior performance across various scales. Addressing the same context, Qin\cite{QIN2022108547} introduced a spatial similarity-based attention mechanism, designed to ensure local intra-level continuity. Similarly, the work in~\cite{MA2022108465} proposed a novel dual attention fusion module. This module was devised to explore feature interdependencies across spatial and channel dimensions, effectively blending features within both missing and known regions. The experiments showed  that, this approach enables the synthesis of smooth content with rich textural details.

\textbf{Semantics embedding}. Semantic information has been increasingly considered in image inpainting in recent years~\cite{zhang2022perceptual}. In~\cite{zhao2020uctgan}, a new cross semantic attention layer was introduced to exploit the long-range dependencies between the known parts and the completed parts, which can improve realism and appearance consistency of repaired samples. Another notable advancement, as proposed by~\cite{zhang2018semantic}, is the Progressive Generative Networks (PGN) framework. This approach addresses semantic image inpainting through a curriculum learning strategy. Furthermore, in~\cite{liao2021image}, a Semantic-Wise Attention Propagation (SWAP) module was developed to enhance the refinement of image textures across different scales. By exploiting non-local semantic coherence, this module effectively minimizes texture mixing artifacts.

\textbf{Transformer Learning}. More recently, transformer-based models utilizing the attention mechanism have been employed for image inpainting~\cite{li2022mat}. ICT\cite{wan2021ict} harnesses the strengths of both CNNs and transformers, utilizing the transformer for reconstructing appearance priors and employing CNNs to replenish textures. Diffusion models have also been applied to image restoration. Repaint~\cite{lugmayr2022repaint} is a DDPM-based approach~\cite{ho2020denoising} that could generate high-quality images, and fill the missing regions. Methods based on diffusion models exhibit high randomness and can generate a wide variety of images. However, in mural inpainting tasks, our emphasis lies more on reality rather than diversity.


\subsection{Mural Inpainting}
Mural image inpainting is a branch of image inpainting, which has attracted more and more attention in recent years~\cite{zou2014chronological,suvorov2022resolution}. At first, image inpainting methods are adopted to restore mural images. In~\cite{jaidilert2018crack}, a semi-automatic scratch detection method was proposed for mural damage localization, based on which the pixel filling and color restoration was performed by different variational inpainting methods. With the success of deep learning techniques in visual tasks, researchers started to apply deep learning-based techniques to handle the mural inpainting problem. An illustrative example is found in~\cite{song2020image}, which achieved much better performance over the traditional methods.

Some methods considered the importance of structure for murals. A structure-guided mural inpainting method~\cite{ciortan2021colour} built on EdgeConnet~\cite{2019EdgeConnect} obtained good results in mural restoration. In~\cite{wang2019inpainting}, line drawings were used to guide the mural inpainting under a sparse-representation framework. In this method, the results obtained are more realistic and are closer to the original mural paintings than that of the conventional methods.

Some methods realized the importance of fidelity in image inpainting~\cite{li2022misf}. A two-phase original-restoration-driven learning method~\cite{wang2020damage} was proposed to guide the model to restore the original content of the Thanka mural. A JPGNet~\cite{guo2021jpgnet} was proposed to combine the advantages of predictive filtering and generative network, which can preserve local structures while removing the artifacts and filling the numerous missing pixels, based on the understanding of the whole scene.

Some methods took the partial convolution in their networks to get better performance. Partial convolutions assisted with the sliding window were employed in~\cite{chen2019image} for mural inpainting. However, the relation of structure retention and color restoration in this method was not well resolved yet. In~\cite{2021Thanka}, a Thanka mural inpainting method was proposed based on multi-scale adaptive partial convolution. On the stroke-like damages, this method obtains results that are relatively close to the real value. While on heavily damaged regions with complex textures, it still has limitations and difficulties in achieving visual plausible results.

\textbf{Summary}. The task of mural inpainting, as a subset of image inpainting, has progressed alongside the evolution of image inpainting techniques. However, the relevant research remains fragmented, lacking a shared dataset and a unified benchmark for direct comparisons. This fragmented landscape leads to isolated efforts among researchers working in the field of mural inpainting, hindering the potential acceleration of research progress through the accumulation of shared knowledge. Therefore, there is an urgent need for a publicly available dataset and a benchmark method to lay the foundation for further in-depth exploration in the realm of mural inpainting. Therefore, this paper undertakes such an endeavor by presenting a public dataset and devising a baseline image inpainting method tailored for murals.

	\section{Methodology}\label{sec:method}
In this section, we first introduce the system overview and the network design, and then describe the self-attention module and loss functions of the proposed neural network.

\subsection{Overview}

\begin{figure*}[!t]
	\centering
	\includegraphics[width=0.9\linewidth]{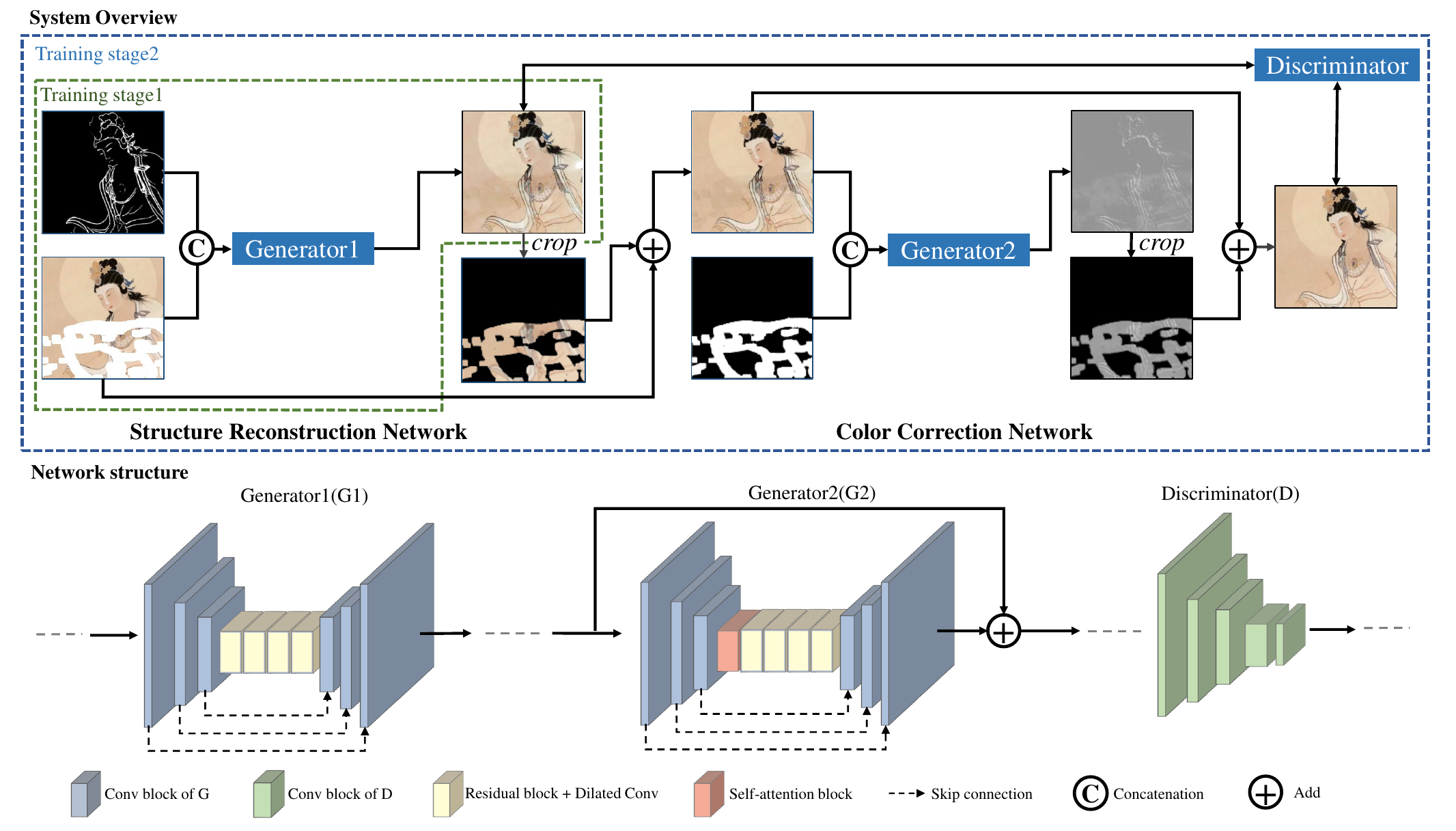}
	\caption{The system overview and network structure of the proposed network. The upper part illustrates the system overview, the lower part illustrates the details of network structure. The whole model is composed of a Structure Reconstruction Network (SRN) and a Color Correction Network (CCN). SRN generates an approximate result with a complete structure, which is input into the CCN to generate a color-consistent final result.}
	\label{fig:networkstructure}
\end{figure*}

In mural inpainting, we adopt line drawings as expert knowledge as they represent the structure of a paint and play a definitive role in terms of semantics. 
Figure~\ref{fig:networkstructure} illustrates our approach. Given a mask image, we combine it with the corresponding line drawing as the input of the structure reconstruction generator (G1). Then, it fills the missing regions, and the generated image is concatenated with the irregular mask as the input of the color correction generator (G2), which will evaluate the residual adjustments to reduce the color bias. Finally, the inpainted image is obtained, where the pixels in the missing regions are adjusted appropriately to the outside pixels.

The input of our network model includes a mural image, a line drawing, and a mask. The mask indicates the missing region to be recovered. In the first stage, we generate an overall well-structured result by the structure reconstruction network (SRN), and then rectify the local hole-region by the color correction network (CCN) in the second stage. Therefore, the masked image and line drawing are the input to SRN, and the mask channel is the input to CCN.

Our inpainting method works from global to local. In SRN, the image generated by G1 directly participates in loss calculation, getting all the pixels backward so that the network is trained to generate globally reasonable results. In CCN, a four-channel image combined by the outcome image of G1 and a mask channel, is fed into G2 to evaluate the residual adjustments, which applied to the missing regions only. The final image combined of the to-be-recovered regions and the background of the original input image, is differentiated by discriminator loss and color correction loss to refine the local results.

Accordingly, the training is divided into two stages. In training stage 1, SRN is trained separately. The training stage 2 does not start until G1 and D come up to converge. Since D has already learned to distinguish the ground truth and the initial generated results, G2 will converge quickly under the guidance of D.


\subsection{Network Architecture}

\noindent {\bf Structure Reconstruction Network (SRN)}. In the structure reconstruction network, we concatenate the damaged mural painting with the corresponding edge map as the input to G1. The network has an encoder-decoder architecture\cite{badrinarayanan2017segnet}. Firstly, through three down-sampling convolutions, the size of the image shrinks to one-eighth of the initial. Then there are three up-sampling convolutions and four residual blocks. We apply instance normalization\cite{ulyanov2016instance} across all layers of the generator. Note that there are skip layers connecting every up-sampled layer with the former corresponding down-sampled layer. Skip layers give the generator a way to circumvent the down-sample bottleneck for information\cite{isola2017image}. Specifically, these skip layers convey low-level multi-scale details for the last few layers to be restored to the original resolution. Moreover, initial color information gets constantly emphasized through skip layers such that the color bias would be enforced to the greatest extent.

\noindent {\bf Color Correction Network (CCN)}. Color correction network (CCN) takes the concatenation of the image and its mask as input, the backbone architecture of CCN keeps the same with SRN. CCN takes a global residual strategy, meaning that the input mural painting of G2 is added to the output of G2, the task of CCN boils down to evaluate  the residual values of the structure-reconstructed image. We find that the global residual contributes a lot to the inpainting effect in our network, as the image has already been well-restored through SRN and what CNN need to do is to slightly adjust the missing-region pixels instead of modifying the entire image. Moreover, a self-attention mechanism is employed in CCN, which  estimates the scores of attention at each pixel. The attention mechanism extends the receptive field to global and exploits the high similarity area to adjust pixel values~\cite{huang2022joint}. Therefore, for blocks with repetitive structure, the network can restore the color consistent with other blocks.

\noindent {\bf Discriminator}. Due to the different training purposes of the two stages, existing two-phase GANs tend to install distinct discriminators for the two generators respectively \cite{2019EdgeConnect,xiong2019foreground}, or just follow the second generator \cite{iizuka2017globally,2019FreeForm}. However, in our network model, both generators share a single discriminator. Although the aims of two networks are different: the former focuses on the integration of the whole structure while the latter focuses on local color difference adjustment, the final training purpose is still the same, that is, to generate images identical to the ground truth. Thus, both generators connect with the same discriminator in our network.

For discriminators, we use the same 70$\times$70 PatchGAN~\cite{2019FreeForm} architecture as in~\cite{2019EdgeConnect}. The 70$\times$70 PatchGAN determines whether or not overlapping image patches of size 70$\times$70 are real. Each green convolution block constitutes discriminator in Figure~\ref{fig:networkstructure} includes a convolution operation, a spectral normalization\cite{miyato2018spectral} and a leakyRelu activation~\cite{maas2013rectifier}.

\subsection{Self-Attention Module}
In CCN, the non-local operation \cite{wang2018nonlocal} is employed. As convolutional operation processes a local neighborhood, either in space or time domain, long-range dependencies can only be captured when these operations are carried out repeatedly, which is computationally inefficient. The non-local operation does not confine to the local neighborhood but computes the response at a position as a weighted sum of the features at all positions in the input feature maps. In CCN, all the known pixels are calculated to optimize the color of pixels inside the hole. For images with large missing area, deep-inside-hole regions require to establish connections with the outside to adjust color, so non-local operation is adopted in our work.

The non-local operation is defined as:

\begin{equation}
	\mathbf{y}_i{}=\frac{1}{\mathcal{C}(\mathbf{x})}\sum_{\forall j}^{}f(\mathbf{x}_i,\mathbf{x}_j)g(\mathbf{x}_j),
\end{equation}
where $i$ is the index of an output position whose response is to be computed and $ j $ is the index that enumerates all possible positions. $ x $ is the input image, and $ y $ is the output image holding the same size as $ x $. $ f $ indicates similarity between position $ i $ and position $ j $ in $ x $, and $ g $ represents the pixel value at the position $ j $. The response is normalized by a factor $  C(\mathbf{x}) $.
Embedded Gaussian is used as $ f $ function, formulated as $ f(\mathbf{x}_i,\mathbf{y}_i)=e^{ \theta(\mathbf{x}_i)^{T}\phi (\mathbf{x}_j) } $. Given $ C(\mathbf{x})=\sum_{\forall j}f(\mathbf{x}_i,\mathbf{y}_i) $, then we get
\begin{equation}
	\mathbf{y}_i=softmax(\theta(\mathbf{x}_i)^{T}\phi (\mathbf{x}_j))g(\mathbf{x}_i),
\end{equation}
where $ g(x)=wx $, $ w $ is a learnable weight matrix. Specifically, it first projects the feature map $ x $ into two spaces, and computes the similarity between every two pixels by dot product, then calculates through a $ softmax $ activation function, and finally computes the attention map by dot product with $ g(\mathbf{x}) $.

\subsection{Loss Functions}

The loss function of GAN consists of two parts: generator loss and discriminator loss\cite{goodfellow2014generative}. Discriminator loss measures whether the discriminator's judgment is consistent with the truth. Adversarial loss in generator loss measures how close the generated picture is to the real one. In the game between these two loss functions, the resultant image will gradually come near to the ground truth.

In our network, the generator loss ($L_{Generator}$) is composed of the perceptual loss~\cite{gatysstyletransfer}, L1 loss, and histogram loss~\cite{risser2017stable}, as defined by Eq.~\eqref{generatorloss}: 

\begin{equation}
	\label{generatorloss}
	L_{Generator}=L_{Perceptual}+L_{1}+0.0005L_{Histogram}.
\end{equation}
The perceptual loss ($L_{Perceptual}$) aims for general image reconstruction, L1 loss ($L_{1}$) for color correction, and histogram loss ($L_{Histogram}$) for rectifying the color distribution while avoiding ambiguity. In the following, we introduce these losses in detail.

\noindent {\bf Perceptual Loss} Perceptual Loss is proposed in style transfer \cite{gatysstyletransfer}, which is widely used in the field of style transfer and image repair in recent years. It contains style loss and content loss, as shown in Eq.~\eqref{gramloss},

\begin{equation}
	\label{gramloss}
	L_{Perceptual}=0.1 L_{Content}+250 L_{Style}.
\end{equation}

Gram matrix can express pixel arrangement features to simulate image style. In mural inpainting, the color arrangement has obvious style characteristics, thus, we apply gram loss in the inpainting task so that the generated part has a similar color style as the known pixels, and make the overall vision be natural and reasonable.

The style loss is fomulated by Eq.~\eqref{styleloss}:
\begin{equation}
	\label{styleloss}
	L_{Style}=\sum_{l=0}^{L}w_l\frac{1}{{4N}_l^2M_l^2}\sum_{i,j}{(G_{ij}^l-A_{ij}^l)}^2,
\end{equation}
where $ G^l $ and  $ A^l $ are respectively the gram matrix of output image and target image at layer $ l $.  $ N_l $ is the number of different feature maps in layer $ l $, and  $ M_l $ is the volume of feature maps in layer $ l $.

The content loss is formulated by Eq.~\eqref{contentloss}:
\begin{equation}
	\label{contentloss}
	L_{Content}=\frac{1}{2}\sum_{i,j}\left(F_{ij}^l-P_{ij}^l\right)^2,
\end{equation}
where $ F^l $ and $ P^l $ are respectively the feature maps of the output image and reference image in layer  $ l $ of the pre-trained VGG19\cite{2014Very} model,  $ i $ and $ j $ denote the size of the feature map in layer $l$.

\noindent {\bf L1 Loss} L1 loss, also known as Mean Absolute Error (MAE)\cite{gatysstyletransfer}, is a loss function used to measure the discrepancy between predicted values and actual values. It calculates the average of the absolute differences between predicted and actual values. Our goal is to minimize the distance during training as much as possible. The L1 loss can be expressed by Eq.~\eqref{l1loss}:

\begin{equation}
	\label{l1loss}
	MAE=\frac{\sum_{i=1}^{N}\left|y_i-y_i^p\right|}{N},
\end{equation}
where $N$ is the number of samples, $y_i^p$ represents the predicted value of the $i_{th}$ sample by the model, $y_i$ represents the actual value of the $i_{th}$ sample.

\noindent {\bf Histogram loss} Histogram loss can increase the stability in texture synthesis~\cite{risser2017stable,gatysstyletransfer}. Gram matrix tends to produce average results, making the results of texture mapping look fuzzy. However, the introduction of histogram loss can guide the pixel distribution of generated results to maintain original variance and standard deviation. In our experiment, we also noticed the problem of the average distribution of pixels caused by Gram matrix. Besides, as is known, L1 will incentivize an averaged grayish color when it is uncertain which color a pixel should take on, according to the research of Luan~\cite{luan2018deep}. The combination of GAN network and L1 Loss can mitigate such problem. In this work, we introduce histogram loss to more vigorously correct the color distribution range and solve the possible fuzzy problem attributed to Gram matrix. The histogram loss is formulated by Eq.~\eqref{histogramloss}:

\begin{equation}
	\label{histogramloss}
	L_{histogram}=\sum_{l=1}^{L}{\gamma_l\left\|O_i-R\left(O_i\right)\right\|_F^l},
\end{equation}
where $ O $ is the output image, $ R\left(O\right) $ is the histogram matching result between $ O $ and the target image, and $ \gamma_l $ is a weight controlling the influence of layer $ l $ .

The loss functions described above are used at different training stages. As mentioned above, in the first training phase, we only train the structure reconstruction network. The aim of this stage is to generate a basic result from the line drawing and damaged image.  As a rectified term, the histogram loss performs better with a preliminary result, and the time complexity and space complexity of the algorithm are both high. Therefore, we do not use the histogram loss in the first stage. In the second training phase, perceptual loss, L1 loss, and histogram loss are used together for the two generators. Since color correction network focuses on local color adjustment, the loss function of G2 only computes the loss in missing regions.

\section{Experiments and Results}\label{sec:exp}
In this section, we introduce the dataset used in the experiments, the training strategy, the comparison with the-state-of-the-art, and the results of ablation study.

\begin{figure}[!t]
	\centering
	\includegraphics[width=0.8\linewidth]{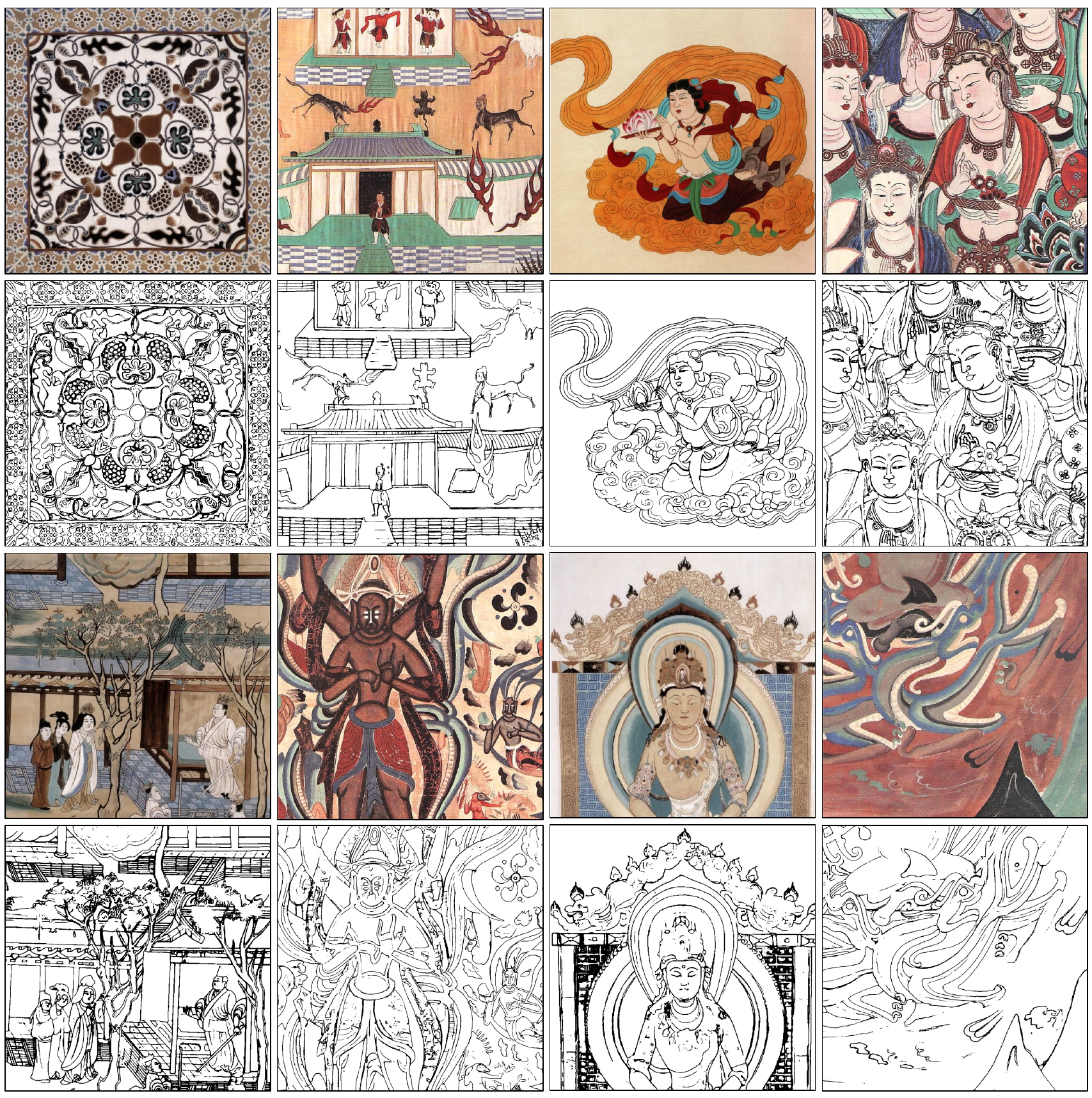}
	\caption{Mural paintings and corresponding line drawings in the DhMurals1714 dataset.}
	\label{fig:dataset}
\end{figure}

\begin{figure*}[!t]
	\centering
	\includegraphics[width=1\linewidth]{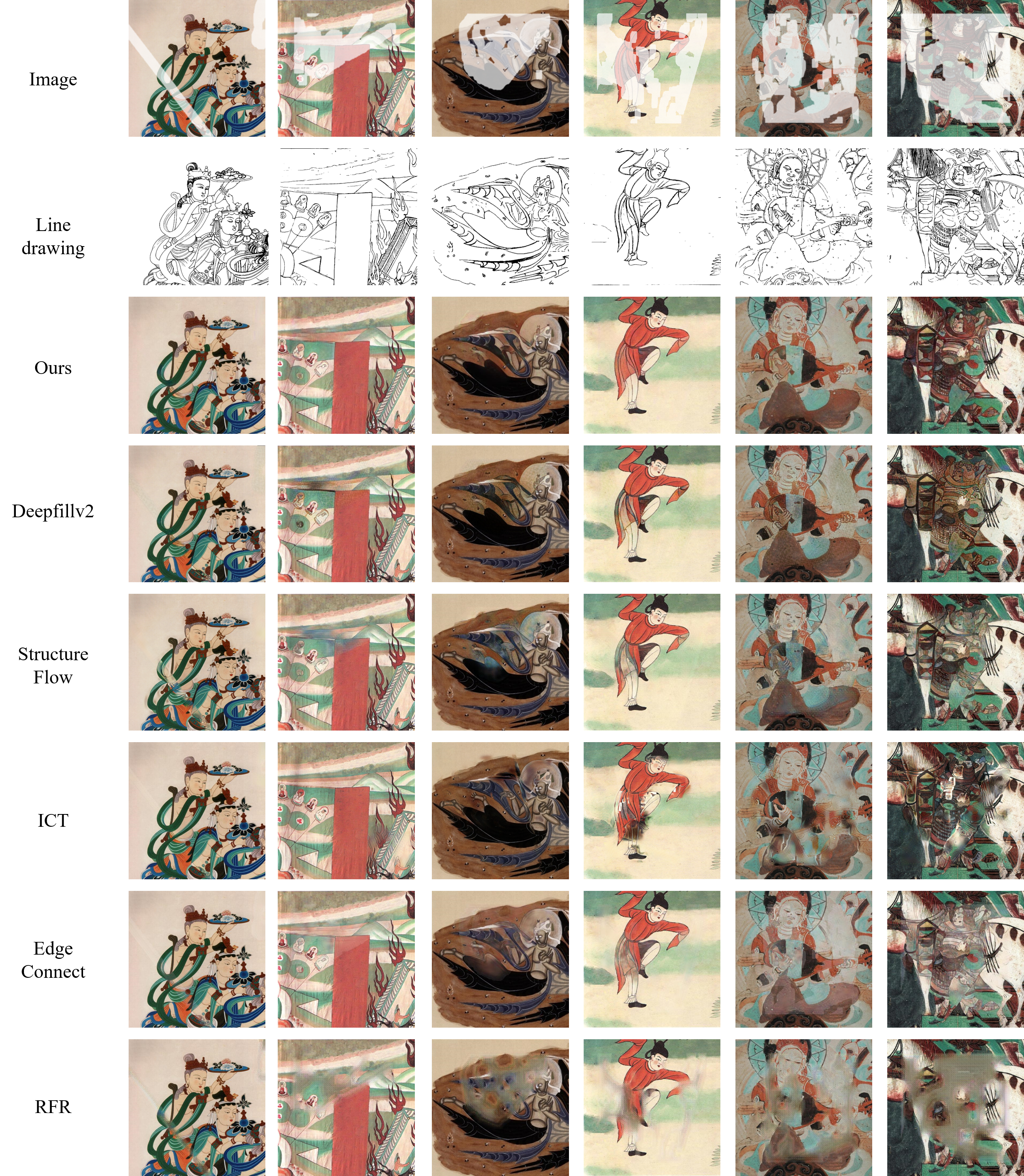}
	\caption{Inpainting results of six mural images obtained by our method and five comparison ones - Deepfillv2~\cite{2019FreeForm}, StructureFlow\cite{ren2019structureflow}, ICT\cite{wan2021ict}, EdgeConnect~\cite{2019EdgeConnect}, and RFR~\cite{li2020recurrent}. The upper row showcases damaged murals with varying mask rates, specifically 10\%, 20\%, 30\%, 40\%, 50\%, and 60\%. The second row displays the corresponding line drawings, followed by six subsequent rows depicting the inpainting outcomes from each method. Columns 1, 2, and 4 represent the replicated murals, while columns 3, 5, and 6 depict real murals from the training set.}
	\label{fig:contrast}
\end{figure*}

\subsection{DhMurals1714 Dataset}
From Dunhuang Mogao Grottoes, due to the extensive damage from nature weathering and human destroy, there are not so many complete or well-preserved mural paintings available for training a deep neural model. In our research, we collect some replicas of murals by artists in addition to the real mural paintings. The real mural paintings are captured by digital cameras, while the replicas of murals are scanned from albums. A number of 1,714 images are collected, including 525 real murals and 1,189 replicas. We randomly selected 50 as the test set, 100 as the validation set, and the remaining 1,564 as the training set. Through data augmentation operations such as rotation, random cropping, color transformation and flipping, a total of 75,072 images constitute our dataset -- DhMurals1714. For masks, we adopt the public mask dataset released in~\cite{pconv}.

In our dataset, line drawings are also contained along with mural images. To making sure that there is no pixel offset between line drawings and original images, line-extracting methods are employed to derive the line drawings. From our tests, DexiNed~\cite{2020Dense} can generate thin line maps that are realistic enough for the human eyes, while being applied without pre-training or fine-tuning, which fits well with our line-extracting task. We employ this method in our data preparation phase. Even though, noise is still unavoidable. In order to denoise to a large extent, we use the bilateral filter~\cite{tomasi1998bilateral} to pre-process the mural image, which can reduce noise and smooth the image while maintaining the structure edges. After that, we extract line maps with the DexiNed model. Finally, a threshold is set to binarize the line maps. Figure~\ref{fig:dataset} shows some of our final line maps. For real mural damages, the missing structures in the line maps are repaired with the help of professional painters.


\subsection{Training Strategy}
Since the training process is divided into two parts, it is inevitable to have cumulative error because the latter generator depends on the output of the former generator. In the early stage of training, the highly randomness of generating results steer the output to be far from the ground truth, then the latter generator takes the wrong images as input so that the final result is severely inaccurate. Training the second generator during these early stages becomes essentially fruitless. Therefore, we took a two-stage training strategy: In the first training phase, SRN gets trained separately. After 8 epochs of training, G1 and D come to converge, G1 starts to generate well-structured paintings, integrating the structure information in edge maps. Then, add CCN to the training in the second phase. On account of the discriminator being well-trained, G2 converges quickly under its guidance.

We implement our model with PyTorch and Cuda. Our network is trained using the ADAM optimizer~\cite{kingma2014adam} with a total of 16 epochs on four NVIDIA GTX 2080 GPUs. The learning rate of D and G are set to $0.0001$ and $0.00001$ respectively, the epoch and batch size are set to $8, 32$ in the first training stage and $8,8$ in the second training stage.

\subsection{Comparisons with State-Of-The-Art Methods}

In this subsection, we compare the proposed method with five advanced methods, including Deepfillv2~\cite{2019FreeForm}, StructureFlow~\cite{ren2019structureflow}, ICT~\cite{wan2021ict}, EdgeConnect~\cite{2019EdgeConnect}, and RFR~\cite{li2020recurrent}.  In qualitative experiments, we analyze the advantages and disadvantages of each method. In quantitative experiments, we measure the results under several different metrics. Then, user study is conducted to compare the performance in another aspect. At last, ablation experiments are used to justify the effectiveness of designed modules.

DeepFillv2, Structure Flow, ICT, and EdgeConnect are trained with edges. RFR is not able to be trained with edges due to its inpainting mechanism.  All of these methods are trained with the DhMurals1714 dataset. We conduct qualitative analysis and quantitative analysis respectively to demonstrate the superiority of our method.



\noindent {\bf Qualitative Comparison.} Figure~\ref{fig:contrast} shows the comparison of our method and three state-of-the-art methods. We observe that, EdgeConnect generates well-structured results owning to the involvement of line drawings, but shows severe color inconsistency between the hole and the background. RFR tends to produce checkerboard artifacts, and the structure can not be restored without the joint of line drawings. DeepFillv2 can restore relatively high-quality images with bright colors and clear textures, but it lacks the comprehensive consideration of overall colors. In comparison, our method achieves complete structure, clear detail and overall consistent colors.

Deepfillv2 is capable of completing structures with high confidence and relatively clear colors under the guidance of line drawings, possibly due to the effectiveness of its multi-scale processing. However, the colors and textures it generates often do not match the background, producing fixed textures regardless of the background, which leads to a lack of harmony between details and the overall image. A similar situation occurs with colors, as some areas exhibit conspicuous differences due to insufficient consideration and integration with the background.

Figure~\ref{fig:contrast} also shows that, EdgeConnect can generate reliable structures within the training framework, however, still have noticeable color discrepancies, where repaired regions are distinctly separated from the background. We also observed similar occurrences in the original EdgeConnect model's experimental results, indicating that the network is not particularly proficient in color processing. This actually became the basis for a targeted improvement in our approach, where we introduced a detail optimization network to achieve finer restoration that can address color discrepancies.

RFR is capable of inferring the general division of color blocks but struggles to deduce intricate details. When the radius of missing information is large, the center of the hole can result in a failed inference, appearing as a blur void. We attribute this to two reasons: Firstly, the muralset dataset is relatively small, with individual images carrying a substantial amount of information and intricate color complexity, making it challenging for the RFR inference module to deduce intricate details. Secondly, the absence of guided line drawings weakens the network's overall structural perception. This also underscores the importance of guided line drawings in aiding the restoration of complex murals.

With the combined assistance of guided line drawings and smoothed images, StructureFlow is capable of restoring reasonable structures. However, it's observable that the repaired regions might exhibit blurriness and color distortion. We believe that this phenomenon arises when the flow direction of feature sampling is not smooth, and as the depth of the hole increases, the difficulty of feature sampling escalates, leading to unclear central regions within the hole.

\begin{figure*}[!t]
	\centering
	\includegraphics[width=0.98\linewidth]{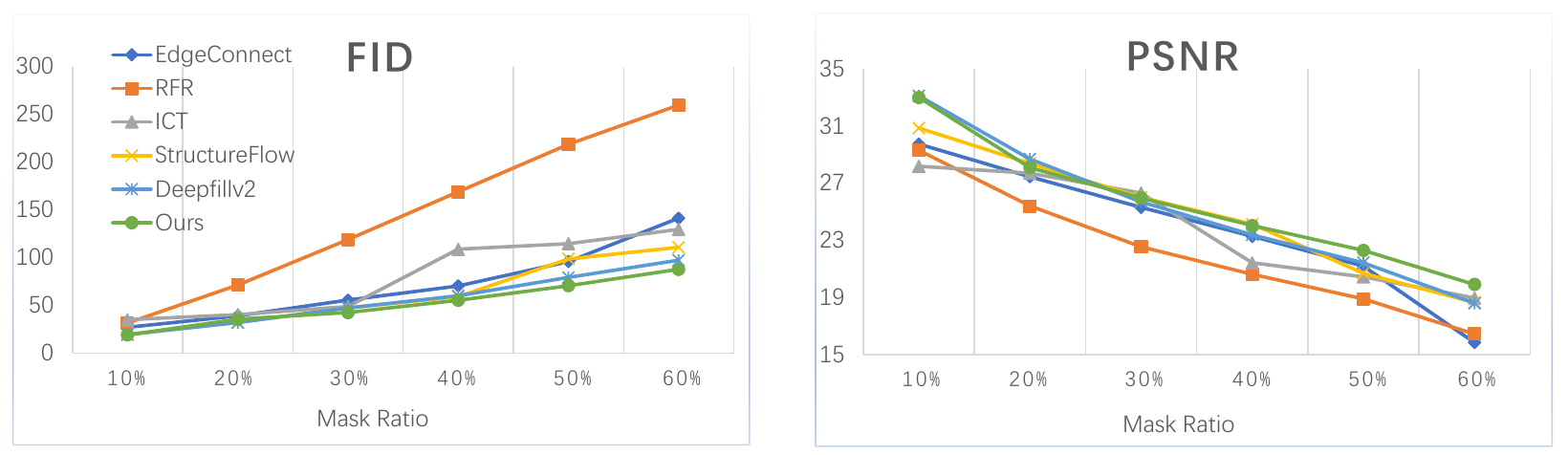}
	\caption{Line charts of FID and PSNR in different ratio of masks. The left chart shows FID values of the six methods in different ratio of masks, the right shows PSNR values of them.}
	\label{fig:linechart}
\end{figure*}

ICT employs a Transformer-based architecture. It can be observed that, in regions with simple and clear structures, ICT is capable of producing sharp and accurate results. However, in regions with complex structural missing, ICT might produce significant artifacts in some areas, resulting in an overall image appearance with ghosting shadows. We attribute this phenomenon to the independent two-stage training: during the global restoration stage, ICT uses downsampled ground truth images as target guidance, leading to results that closely match the downsampled ground truth. Due to the training discontinuity, the second stage, trained independently, does not use the results of the first stage as input. Instead, it employs downsampled ground truth images and line drawings as input guidance to restore the image. Consequently, when the results from the first stage do not perfectly align with the ground truth structure, disarray and ghosting effects occur during inference.

We also conduct restoration experiments without the assistance of line drawings. In this scenario, it is very challenging to achieve high-resolution image restoration without the aid of line drawings. In most cases, the results suffer from severe image artifacts. These findings indicate that, the Transformer architecture possesses strong global perception capabilities since it can generate sharp and accurate results when the structure is aligned with ground truth. However, integrating the Transformer with structural information remains a challenge to be addressed. Additionally, the Transformer places high demands on hardware, presents significant training challenges, and exhibits slower inference speeds compared to other methods,as observed in Table~\ref{Computation}. Therefore, how to apply the Transformer in image inpainting in a lightweight and more reasonable manner remains a topic that requires further exploration.

\begin{table}[!t]
	\begin{center}
		\caption{Quantitative results obtained by six models: Ours, Deepfillv2, StructureFlow, ICT, EdgeConnect, and RFR. The best result of each metric is marked in bold font.}
		\label{metrics}
		\resizebox{0.7\textwidth}{!}{
			
			\begin{tabular}[1]{@{}lccccc}
				\toprule
				Method & SSIM$\uparrow$ & MSE$\downarrow$ &PSNR$\uparrow$ & LPIPS$\downarrow$ &FID$\downarrow$\\
				\midrule
				Ours & {\bf 0.9003}&{\bf 0.0023}&{\bf 27.1254}& {\bf 0.0503}& {\bf 48.46}\\
				DeepFillv2\cite{2019FreeForm} & 0.8992&0.0027&26.5359&0.0547&49.18\\
				EdgeConnect\cite{2019EdgeConnect}& 0.895&0.0027&26.1661&0.0575&55.42\\
				RFR\cite{li2020recurrent}& 0.8311&0.0053& 23.2756&0.1241&118.8\\
				StructureFlow\cite{ren2019structureflow}& {\bf0.9003}&0.0026&25.9084&0.0574&49.25\\
				ICT\cite{wan2021ict}& 0.829&0.0076&21.9816&0.1277&100.9\\
				\bottomrule
			\end{tabular}
		}
		
	\end{center}
\end{table}


\noindent {\bf Quantitative Comparison.} Several metrics are adopted to illustrate the difference between these methods. 
Specifically, we calculated SSIM\cite{2004Image}, MSE, PSNR\cite{hore2010image}, LPIPS\cite{zhang2018unreasonable}, and FID\cite{heusel2017gans} as evaluation metrics. MSE measures pixel-level differences, SSIM quantifies structural similarity, PSNR indicates image quality, LPIPS evaluates image similarity perceptual to humans~\cite{xiang2023deep}, and FID calculates the distance between feature vectors of real and generated images. These metrics are computed based on a validation set of 100 mural images. Each image was randomly assigned an irregular mask. As shown in Table~\ref{metrics}, our method achieves outstanding results, with the highest SSIM and PSNR values, and the lowest MSE, LPIPS, and FID values.


\begin{table}[!t]
	\begin{center}
		\caption{Computational Complexity Analysis. DF refers to DeepFillv2, EC refers to EdgeConnect, SF refers to StructureFlow.}
		\label{Computation}
		\resizebox{0.7\textwidth}{!}{
			\begin{tabular}[1]{lllllll}
				\toprule
				& Ours &DF &EC &  RFR&SF& ICT\\
				\midrule
				FLOPs(G)   & 7.35    & 31.18  & 24.65  & 82.47 & 105.17 & 46264.96\\
				\#Parameter(M)   & 11.90  & 15.35 & 10.77  & 30.59 & 92.53  & 441.39\\
				Infer. time(ms)  & 22 & 109   & 18  & 97 & 43 & 15280 \\
				\bottomrule
			\end{tabular}
		}
	\end{center}
\end{table}

\begin{figure*}[!t]
	\centering
	\includegraphics[width=1\linewidth]{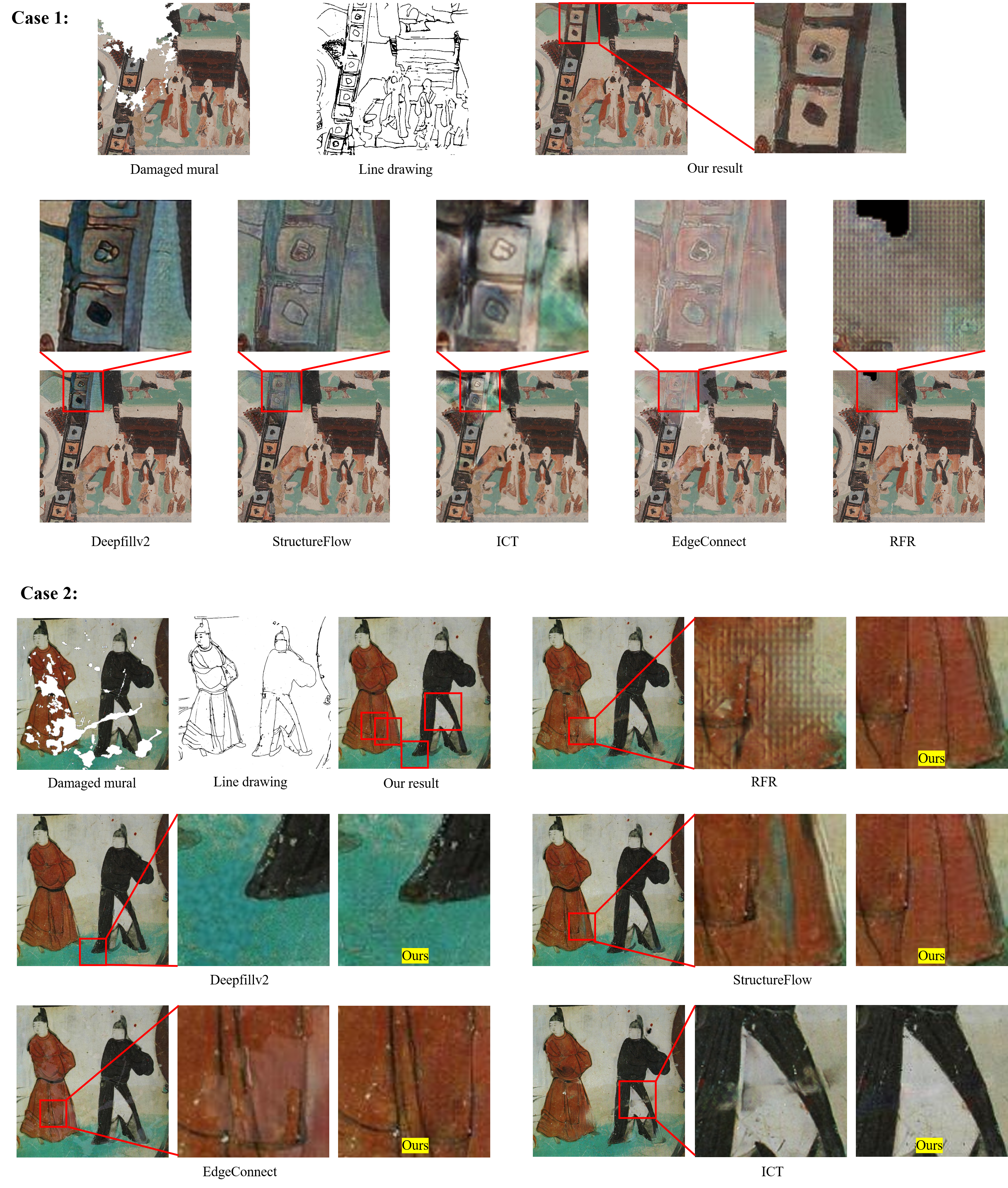}
	\caption{Experimental results obtained by the proposed method and five comparison methods in repairing real damages of mural images.
		In Case 1 and Case 2, the red boxes highlight the inpainting details.}
	\label{fig:realmural}
\end{figure*}

Additionally, we have also compared the performance of these methods under various mask ratios, i.e., 10\%, 20\%, 30\%, 40\%, 50\%, and 60\%, as shown in Figure~\ref{fig:linechart}. We present the curves of PSNR and FID, because PSNR assesses the image quality, FID measures the similarity to ground truth. It can be observed from Figure~\ref{fig:linechart}, the results of PSNR and FID comply with the objective values of mural restoration. 

It can be observed that, at lower mask ratios, the metric values of all methods are relatively close, and as the mask ratio increases, the difference becomes more and more apparent. In our method, the trends of FID increase and PSNR decrease are the most gradual, consistently exhibiting outstanding performance. EdgeConnect and RFR maintain a certain gap with our method throughout the range. Deepfillv2 and StructureFlow initially closely resemble our method, however, as the mask area increases, significant gaps in PSNR and FID become evident. ICT's performance is relatively unstable with significant fluctuations. As explained earlier, due to the independent training of ICT, when the predictions from the first phase do not correspond to the true line drawings, the results appear chaotic; conversely, when they do correspond, the results are clear and accurate. Therefore, ICT exhibits highly unstable behavior. As evident in Table~\ref{metrics}, its MAE value is also the highest among all methods.

We also test the computational complexity of each method, and listed the results in Table~\ref{Computation}. The parameter number and inference time of our method are only slightly larger than EdgeConnect, while FLOPs is the lowest. ICT exhibited the lowest efficiency. This indicates that our approach not only enhances restoration results but also maintains lightweight characteristics. It has potential to be deployed in devices with limited computation resources, such as embedded devices.

\subsection{Experiment on real damaged murals}

In the previous part, we conduct training and test on DhMurals1714 dataset. In order to verify the generality of our method, we conduct test experiment on the real damage murals. 


Figure~\ref{fig:realmural} presents two examples of real damaged mural inpainting. As there are no ground-truth images for damaged murals, we seek the help from expert painters in Dunhuang murals who manually outline the damaged areas and complete the corresponding line drawings. We zoom in on the details to compare the characteristics and differences of the methods. 

In Case 1, where the damaged area is relatively concentrated, we compare the details of the same region. It can be observed that, Deepfillv2 exhibits unrealistic colors, ICT shows artifacts, EdgeConnect has noticeable color discrepancies, and RFR fails to restore larger holes. This complies with the observations we made in our test set. In Case 2, where the damage is more dispersed, we magnify different areas of the restoration results from our method for comparison, and similar results to those previously discussed are obtained. The restoration effect of Deepfillv2 is close to that of our method, but upon closer inspection, its restored details are not as clear as ours.



\begin{figure}[!t]
	\centering
	\includegraphics[width=0.9\linewidth]{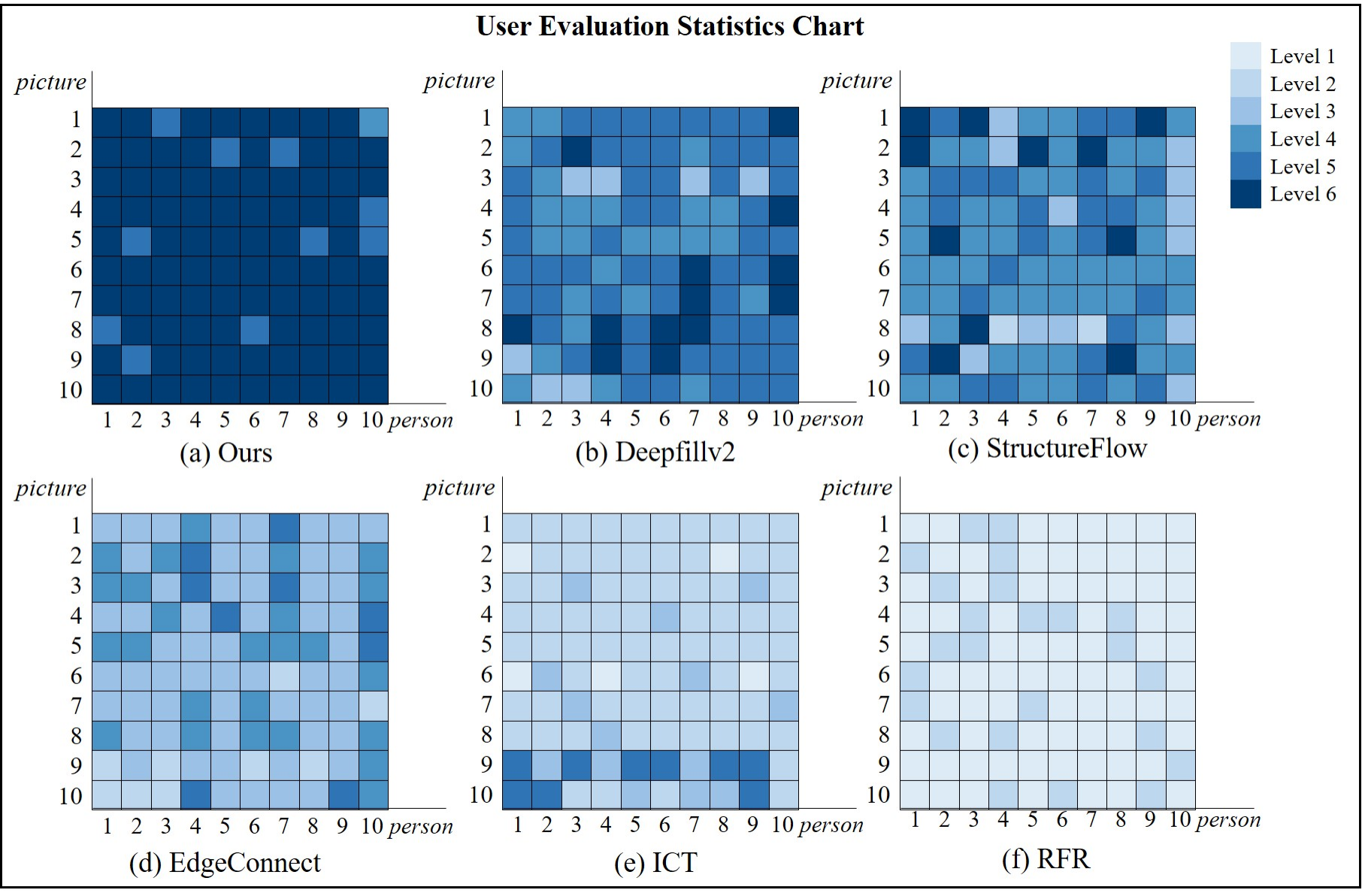}
	\caption{User evaluation statics chart. The six statistical graphs from (a) to (f)  show the user evaluation results of the six methods, respectively. We sorted users' scores from low to high into six level. The higher the score, the higher the level and the darker the color. Note that the same level is allowed.}
	\label{fig:userstudy}
\end{figure}

\begin{figure}[!t]
	\centering
	\includegraphics[width=0.8\linewidth]{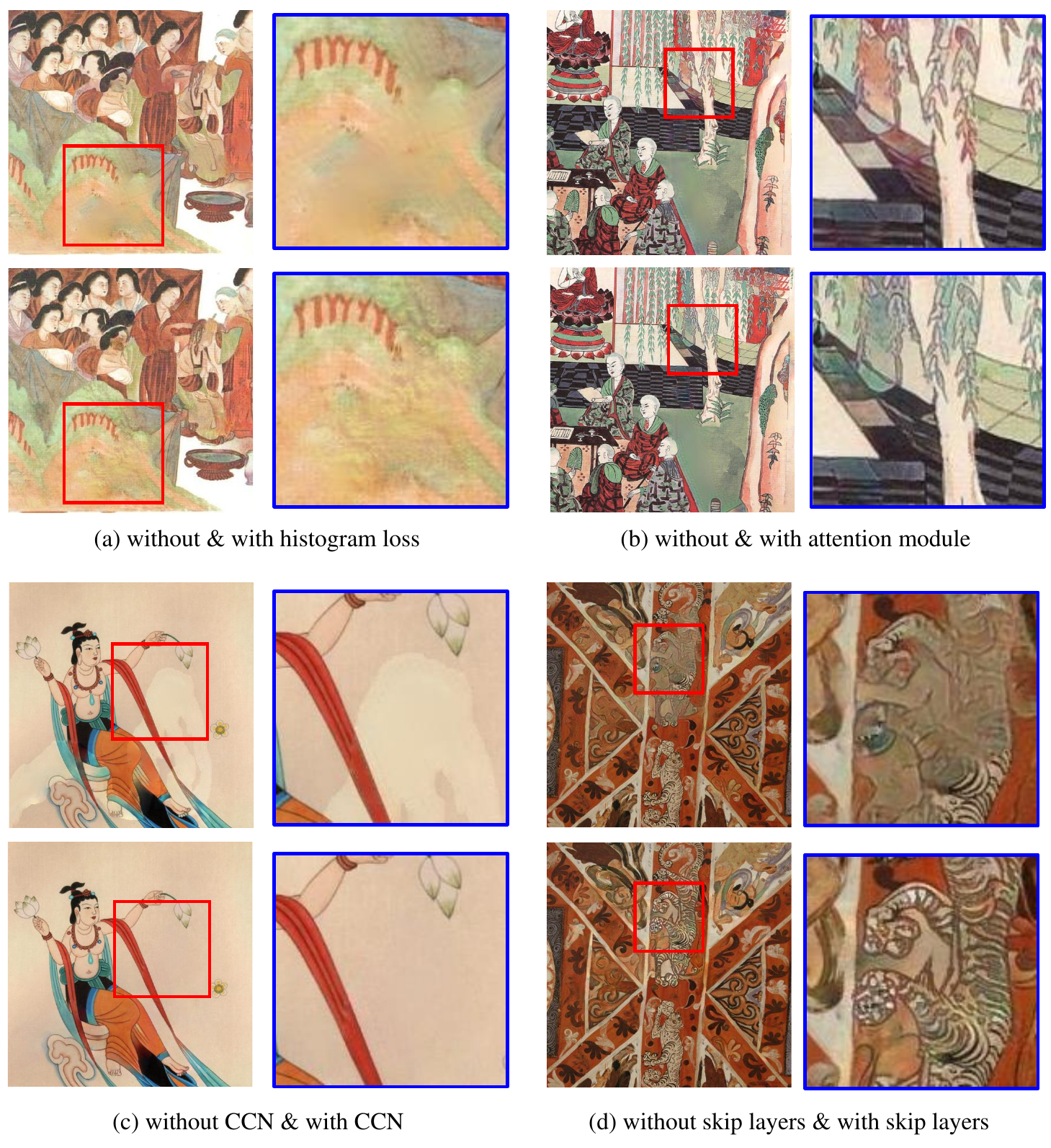}
	\caption{Four samples (a) to (d) were used to show the ablation experimental results of four modules: histogram loss, attention, CCN, and skip layers, respectively. In each sample, the upper image shows the result without a module. In order to demonstrate the results more clearly, we enlarge a part of the image in each ablation experiment.}
	\label{fig:ablation}
\end{figure}

\begin{table}[!t]
	\begin{center}
		\caption{Quantitative evaluations for ablation study.}
		\label{trainingstage}
		\resizebox{0.7\textwidth}{!}{
		\begin{tabular}[1]{@{}lccccc}
			\toprule
			training strategy & SSIM$\uparrow$ & MSE$\downarrow$ &PSNR$\uparrow$ & LPIPS$\downarrow$ &FID$\downarrow$\\
			\midrule
			Whole network & {\bf 0.9003}&{\bf 0.0023}&{\bf 27.1254}& {\bf 0.0503}& {\bf 48.4600}\\
			One-stage training &0.8461	&0.0033	&25.5454&0.0903	&53.3971\\
			Without HL &0.8497&0.0029&26.2559&0.0838&49.7477\\
			Without CCN &0.8198&0.0055&23.1617&0.1158&63.7995\\
			Without attention &0.8480&0.0036&25.2767&0.0908&49.5355\\
			Without skiplayer &0.8253&0.0042&24.5608&0.1026&54.5291\\
			\bottomrule
		\end{tabular}}
	\end{center} \label{tbl:ablation}
\end{table}

\subsection{User Study}
We perform a user study towards the inpainting result of the six methods. We randomly select 10 images from the evaluation set, listed six different restored results with these methods for each image, and ask 10 users to rate the quality of them. Users score them on a scale of 1 to 10 referring to the original paintings we provide, where higher score denotes higher quality. According to the experimental results, our methods got an average score of 8.65, while the scores of Deepfillv2, structureflow, ICT, EdgeConnect, and RFR are 7.85, 7.37, 5.7, 6.82, and 3.93, respectively. In Figure~\ref{fig:userstudy}, six statistical charts from (a) to (f)  show the user-evaluation results of the six methods, respectively. We sorted users' scores from high to low into six levels. A higher score denotes a higher level, marked with darker color. The results justify our analysis that our method achieves the best visual effects as judged by users. Following this, Deepfillv2 and structureflow perform similarly, however they obtain worse results for larger damages, and occasionally show noticeable color inconsistencies, leading to lower scores as compared to our method.


\subsection{Ablation Study}
This section conducts an ablation study to validate the significance of components in our model.
To be specially, CCN, histogram loss (HL), attention mechanism and skip layers (SL) are the four components.

We remove HL, CCN, Attention, and Skip Layers from our model, respectively. 
Experiments in Figure~\ref{fig:ablation} show the results. The left side of each group shows the results of method without a certain module. In order to demonstrate the results more clearly, we select the most typical image in each ablation experiment.
According to Figure~\ref{fig:ablation}, each module is playing an important role in our network:
HL improved the definition of pixels in the hole region; 
the attention module kept the structural and texture consistency; 
CCN rectified obvious color bias in overall; 
skip layers helped to maintain the structural and color consistency. Values in Table~\ref{tbl:ablation} further validate the above results and analysis.

\section{Conclusion}\label{sec:conc}
In this work, we proposed a line-drawing-guided progressive inpainting method for repairing mural damages. The pipeline of inpainting was divided into two steps, i.e., structure reconstruction and color correction, which were implemented by a structure reconstruction network (SRN) and a color correction network (CCN), respectively.  In the experiments, a dataset for mural image inpainting was constructed on 1,714 mural images which were collected from Dunhuang Mogao Grottoes. The proposed method was evaluated against the current state-of-the-art methods. Visual results showed that, the line drawing provided a great help in reconstructing the real structure of severely damaged murals, and the strategy separating the structure restoration and color correction largely alleviated the problem of color bias. Quantitative results also demonstrated the superiority of the proposed method over the competitors. 



\bibliographystyle{elsarticle-harv} 
\bibliography{ref} 


\end{document}